\documentclass[letterpaper, 10 pt, conference]{ieeeconf}  %

\IEEEoverridecommandlockouts                              %

\usepackage{graphicx} %
\usepackage{epsfig} %
\usepackage{mathptmx} %
\usepackage{times} %
\usepackage{amsmath} %
\usepackage{amssymb}  %
\usepackage{booktabs}
\usepackage{gensymb}
\usepackage{cite}
\usepackage{caption}
\usepackage{subcaption}
\usepackage{multirow}
\usepackage{tabularx}
\usepackage[pagebackref,breaklinks,colorlinks]{hyperref}

\usepackage[capitalize]{cleveref}
\crefname{section}{Sec.}{Secs.}
\Crefname{section}{Section}{Sections}
\Crefname{table}{Table}{Tables}
\crefname{table}{Tab.}{Tabs.}

\title{\LARGE \bf
LiDAR View Synthesis for Robust Vehicle Navigation Without Expert Labels
}

\author{Jonathan Schmidt$^{1,2}$ \and Qadeer Khan$^{1,2}$ \and Daniel Cremers$^{1,2,3}$%
\thanks{$^{1}$ Computer Vision Group, School of Computation, Information and Technology, Technical University of Munich. Contact: {\tt\small \{jonathan.schmidt,qadeer.khan, cremers\}@tum.de}}
\thanks{$^{2}$ Munich Center for Machine Learning (MCML).}
\thanks{$^{3}$  University of Oxford.}
}

\begin{document}

\maketitle
\thispagestyle{empty}
\pagestyle{empty}

\begin{abstract}

Deep learning models for self-driving cars require a diverse training dataset to manage critical driving scenarios on public roads safely. This includes having data from divergent trajectories, such as the oncoming traffic lane or sidewalks. Such data would be too dangerous to collect in the real world. Data augmentation approaches have been proposed to tackle this issue using RGB images. However, solutions based on LiDAR sensors are scarce. Therefore, we propose synthesizing additional LiDAR point clouds from novel viewpoints without physically driving at dangerous positions. The LiDAR view synthesis is done using mesh reconstruction and ray casting. We train a deep learning model, which takes a LiDAR scan as input and predicts the future trajectory as output. A waypoint controller is then applied to this predicted trajectory to determine the throttle and steering labels of the ego-vehicle. Our method neither requires expert driving labels for the original nor the synthesized LiDAR sequence. Instead, we infer labels from LiDAR odometry. We demonstrate the effectiveness of our approach in a comprehensive online evaluation and with a comparison to concurrent work. Our results show the importance of synthesizing additional LiDAR point clouds, particularly in terms of model robustness. Code and supplementary visualizations are available at: \href{https://jonathsch.github.io/lidar-synthesis/}{https://jonathsch.github.io/lidar-synthesis/}

\end{abstract}

\section{INTRODUCTION}
\label{sec:intro}

Recent advances in deep learning and computer vision have led to significant progress towards end-to-end autonomous driving solutions, where the model learns to map from raw sensor input to control commands~\cite{chitta2022transfuser, Chitta2021neat, bojarski2016end, pccontrol_2022, prakash2021multi, control-across-weathers-19}. Training data for such models is typically collected by recording the raw sensor data and the corresponding control parameters of vehicles maneuvered by an expert driver~\cite{chitta2022transfuser, prakash2021multi, bojarski2016end}. However, deep learning models tend to fail at inference time when encountering samples that differ too much from the training data. In order to make the trained model robust to deviations away from the road, we also require training data at off-trajectory positions, lateral to the correct driving trajectory.

Collecting such additional off-trajectory data at lateral offsets to the original driving path would require dangerous and illegal maneuvers such as driving in the lane of the oncoming traffic or even sidewalks. This, however, would be intractable. Therefore, we propose using only the original driving trajectory to synthesize additional off-trajectory data at lateral positions. This task is referred to as novel-view synthesis and has recently piqued the interest of the academic community, particularly on image data obtained from RGB sensors~\cite{mildenhall2021nerf, tancik2022block, nndriving2023, wimbauer2023behind}. But what about other sensors?

In addition to cameras, the autonomous driving sensor suite comprises various other sensors such as LiDAR, IMU, and GPS. Among them, RGB cameras and LiDAR sensors have proven to be the most pragmatic combination since they provide a rich representation of the driving environment~\cite{chitta2022transfuser, prakash2021multi, patel2017sensor, patel2019deep, sobh2018end, shao2023safety}. RGB cameras provide dense, 2D information up to pixel accuracy. However, deep learning models trained on image data are susceptible to weather or lighting changes~\cite{sobh2018end}. 
In contrast, LiDAR sensors rely on active depth sensing to create a 3D representation of the ego vehicle's environment\cite{sobh2018end, shao2023safety}. LiDARs are also less influenced by lighting and weather variations~\cite{sobh2018end}. However, unlike images, LiDAR point clouds are sparse. Due to these complementary traits of RGB and LiDAR, plenty of concurrent research has attempted to exploit the best characteristics of both these sensors for the task of vehicle control~\cite{prakash2021multi, chitta2022transfuser, patel2017sensor, patel2019deep, sobh2018end, shao2023safety}. Meanwhile, some recent works have also explored the possibility of only utilizing RGB to train models emulating depth information to improve the performance of vehicle control methods~\cite{pccontrol_2022, nndriving2023}. However, research focusing exclusively on the performance of LiDAR on the task of vehicle control is scarce and is, therefore, the emphasis of this paper.

\begin{figure*}
    \centering
    \includegraphics[width=0.9\textwidth]{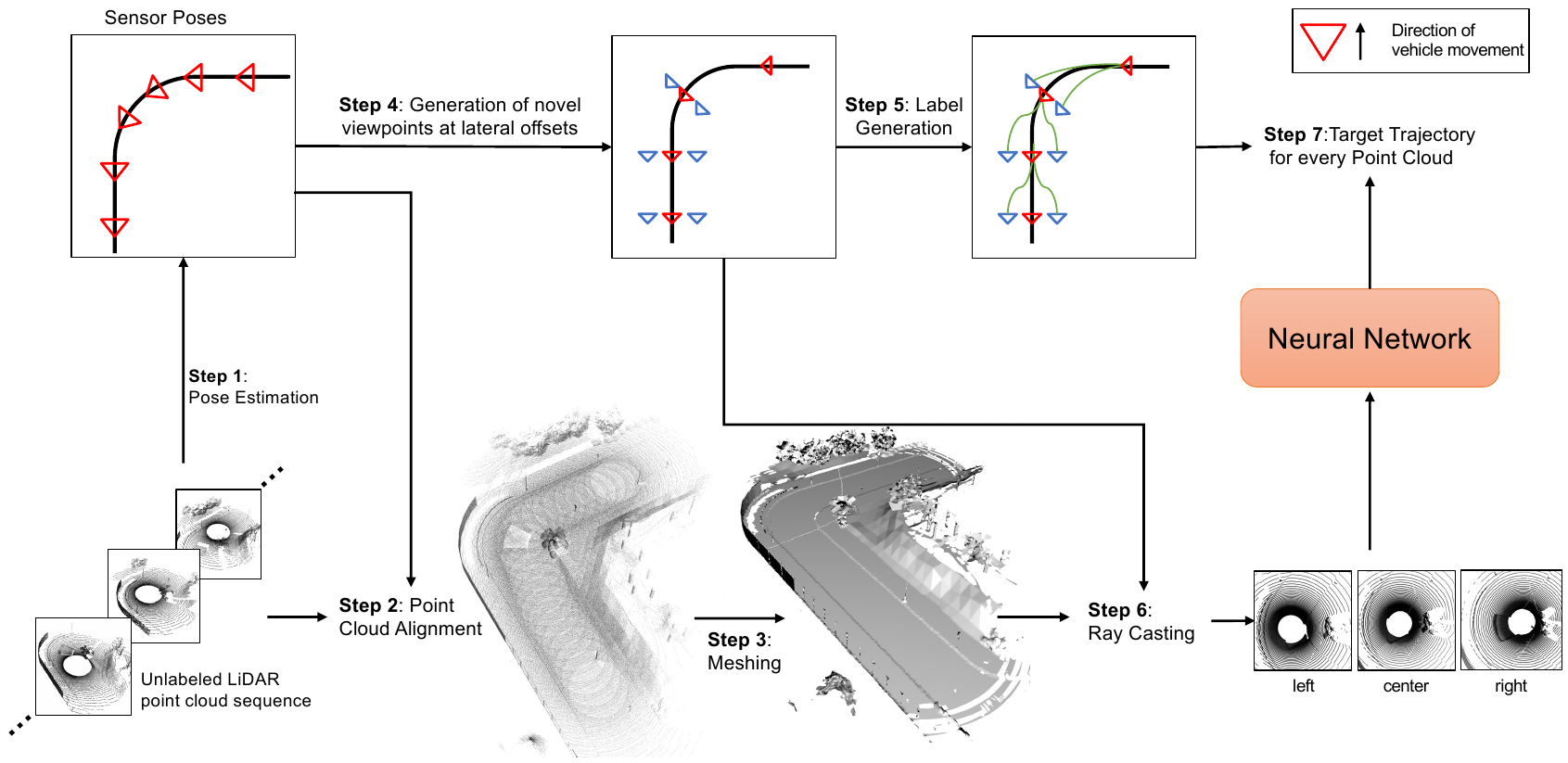}
    \caption{Overview of the framework describing the steps for synthesizing additional LiDAR scans and generating their corresponding labels, given an unlabeled sequence of LiDAR scans from a LiDAR sensor attached to the car. \textbf{Step 1:} (\cref{sec:pose-estimation})  From this sequence of LiDAR point clouds, we compute the relative sensor poses and corresponding trajectory traversed by the vehicle using a LiDAR odometry algorithm. \textbf{Step 2:} (\cref{sec:alignment}) We use these poses to align all LiDAR point cloud scans. \textbf{Step 3:} (\cref{sec:meshing}) Since the aligned point clouds are sparse, we reconstruct a triangle mesh. \textbf{Step 4:} (\cref{sec:viewpoint-gen}) We sample positions at lateral offsets of the original/reference trajectory from where we desire to generate additional off-trajectory data. \textbf{Step 5:} (\cref{sec:label-generation})  The sensor poses determined in Step 2 are used to compute target label trajectories for both the original and lateral offset positions using cubic spline interpolation.  \textbf{Step 6:} (\cref{sec:raycasting}) We use the triangle mesh generated in Step 3 to synthesize additional LiDAR-like point clouds using ray casting at the lateral positions determined in Step 4. \textbf{Step 7:} (\cref{sec:driving-model}) Finally, we can train a neural network which takes both the original LiDAR scans and those created in Step 6, to predict the target trajectory determined in Step 5. Steps 1-6 are only used for training and are not required at inference time. Only the trained neural network is needed during inference.}
    \label{fig:method-overview}
\end{figure*}

More specifically, we address the problem of synthesizing novel LiDAR point clouds from only the available sequence of LiDAR scans. A deep learning model is then trained with this additional data to predict the desired future trajectory, which we convert to vehicle control commands using a waypoint controller. The labels for training the model are implicitly calculated by running an odometry algorithm on the unlabeled sequence of LiDAR scans. This alleviates the problem of requiring an expert driver to collect driving labels at the synthesized viewpoints, which may be located in the lane of oncoming traffic. Figure \ref{fig:method-overview} gives an overview of our framework where a sequence of LiDAR scans are first aligned and then converted into a triangle mesh. Given this triangle mesh, new LiDAR-like point clouds are synthesized at any desired position within a certain range from the original trajectory.

We summarize the contributions of our framework below:
\begin{itemize}
    \item Synthesis of additional off-trajectory LiDAR point clouds, based on mesh reconstruction and ray-casting. This is done using only a single sequence of on-trajectory LiDAR data. This additional data improves online driving performance.
    \item We do not require supervised expert labels for training a  vehicle control model. Instead, we infer labels from LiDAR odometry. Therefore, it is not required to drive in dangerous situations to collect expert driver labels.
    \item We provide the code and additional visualizations of our framework here: \href{https://jonathsch.github.io/lidar-synthesis/}{https://jonathsch.github.io/lidar-synthesis/}.
 \end{itemize}

\section{RELATED WORK}
\label{sec:related-work}

\textit{Data augmentation} in the context of LiDAR point clouds has mainly focused on improving 3D object detection~\cite{hahner2020quantifying, yan2018second, Fang2021LiDAR-aug}. Trivial methods include scaling, flipping, rotation, point jittering, and random dropping of points ~\cite{hahner2020quantifying}. \cite{yan2018second} and \cite{Fang2021LiDAR-aug} place novel objects in the point cloud, which are sampled from a database. \cite{chen2020pointmixup} proposes an interpolation scheme to generate novel point clouds along the trajectory while we generate point clouds at off-trajectory locations. \cite{manivasagam2020LiDARsim} reconstructs road scenarios from multiple LiDAR sweeps and uses additional sensors such as GPS, IMU, and wheel odometry data for vehicle perception. Our method focuses on vehicle navigation and only requires a single run with a LiDAR sensor.

\textit{Deep learning-based autonomous driving} research can be grouped by the sensor signal(s) used as input. The most prominent variants are image-only~\cite{chen2020learning, Chitta2021neat, codevilla2019exploring, behl2020label, ohn2020learning, prakash2020exploring, toromanoff2020end, nndriving2023} and multi-modal~\cite{prakash2021multi, chitta2022transfuser, patel2017sensor, patel2019deep, sobh2018end, shao2023safety, liu2021efficient} approaches, while architectures using only LiDAR are scarce. An early image-based approach was~\cite{bojarski2016end}, which learned to predict steering angles directly from the images of multiple front-facing RGB cameras using recorded control commands from an expert driver. A recent method is Transfuser~\cite{prakash2021multi, chitta2022transfuser}, which uses Transformers~\cite{vaswani2017attention} to fuse RGB and LiDAR data into an implicit feature vector and then applies an autoregressive model to predict a set of waypoints the vehicle should follow. In contrast to our method, Transfuser requires ground-truth training data about the trajectories and relies on an additional RGB camera. Neural attention fields (NEAT)~\cite{Chitta2021neat} encode RGB images into a neural representation that is subsequently decoded into a semantic birds-eye-view image and waypoint offsets, from which the target trajectory is inferred. However, NEAT requires semantic annotations of its environment for training. Our method does not require any specialized data and can be trained solely from an unlabeled sequence of LiDAR scans.

\textit{Novel-View Synthesis} for images is a well-explored research topic. It deals with the problem of generating images from new viewpoints given a single or multiple images of a scene. The majority of recent work uses a variant of neural radiance fields (NeRF)~\cite{mildenhall2021nerf} such as Block-NeRF~\cite{tancik2022block}, which scales the NeRF representation up to city scale. \cite{nndriving2023}~applies Novel-View Synthesis to autonomous driving using monocular depth estimation~\cite{godard2019digging}. While prevalent for images, Novel-View Synthesis for point clouds is scarce, which is why we synthesize novel LiDAR-like point clouds from a sequence of LiDAR scans.

\section{METHOD}
\label{sec:method}

In this section, we describe our proposed framework. Given only unlabeled sequences of LiDAR point clouds, our goal is to train a deep neural network for vehicle navigation/control. The trained model takes a raw LiDAR point cloud as input and predicts the target trajectory from which the control parameters of the ego vehicle are determined using a waypoint-based controller. The framework is illustrated in Figure \ref{fig:method-overview} and can be decomposed into the following components:

\begin{enumerate}
    \item \textbf{Pose Estimation} (\cref{sec:pose-estimation}) computes the relative transformations between the LiDAR point clouds and, thus, the vehicle trajectory which we refer to as the \textit{reference} trajectory.
    \item \textbf{Point Cloud Alignment:} (\cref{sec:alignment}) Given this \textit{reference} trajectory, we merge the point clouds in a common reference frame.
    \item \textbf{Meshing:} (\cref{sec:meshing}) We obtain a tight triangle mesh by applying a surface reconstruction algorithm.
    \item \textbf{Novel-Viewpoint Position Selection:} (\cref{sec:viewpoint-gen}) Using the sensor poses, we select sensor locations at lateral offsets to the \textit{reference} trajectory at which novel LiDAR scans are desired to be synthesized.
    \item \textbf{Label Generation:} (\cref{sec:label-generation}) Using only the estimated sensor poses from step 2, we compute target trajectories for every viewpoint including those generated in step 4.
    \item \textbf{Ray Casting:} (\cref{sec:raycasting}) We use ray casting to synthesize novel, LiDAR-like point clouds from the locations generated in step 4
    \item \textbf{Neural Network Training} (\cref{sec:driving-model}) We train a neural network for vehicle control using the point clouds from step 6 and the target trajectories from step~5.
\end{enumerate}

An advantage of this framework is that individual components are independent of one another. Each component can thus easily be substituted with better solutions if they become available in the future. In the following, we explain these components in more detail.

Our method expects a sequence of $N$ point clouds at timesteps $i \in \{0,..., N-1\}$, each of the form $\mathbf{P}_i = \{\mathbf{p}_j | \mathbf{p}_j \in \mathbb{R}^{3}, j \in \{0,..., L-1\} \}$, in the coordinate system of the LiDAR sensor. $L$ is the number of points in a point cloud. For the remainder of this section, we define transformations as matrices of the form $\mathbf{T}_i \in SE(3)$ w.r.t. to a common global coordinate system, where $SE(3)$ denotes the Special Euclidean group. $\mathbf{T}_i$ refers to the \textit{world-to-camera} transformation. This transformation maps a point from global to local coordinates. We assume: (1) the point clouds in each sequence have some overlap between them, (2) there are no transient objects, and (3) the point clouds in each sequence were captured at a constant frequency. All 3D coordinates refer to the left-handed coordinate system, where the x-axis points forwards, the y-axis to the right, and the z-axis upwards.

\subsection{Pose Estimation}
\label{sec:pose-estimation}

We are initially given an unlabeled sequence of LiDAR point cloud scans. The first step is to compute the relative transformations between the point clouds. This can be done using the Iterative Closest Points (ICP) algorithm~\cite{icp}. ICP provides us with the transformation $\mathbf{T}_i$ from the first point cloud $\mathbf{P}_0$ to the i-th point cloud $\mathbf{P}_i$. Consequently, the first point cloud $\mathbf{P}_0$ is the reference frame, and its pose $\mathbf{T}_0$ is the identity matrix.

\subsection{Alignment}
\label{sec:alignment}

\begin{figure*}[!htb]
    \centering
    \begin{tabular}[t]{ccc}
        \begin{tabular}{c}
            \begin{subfigure}{0.4\textwidth}
            \centering
            \smallskip
            \includegraphics[width=0.9\linewidth]{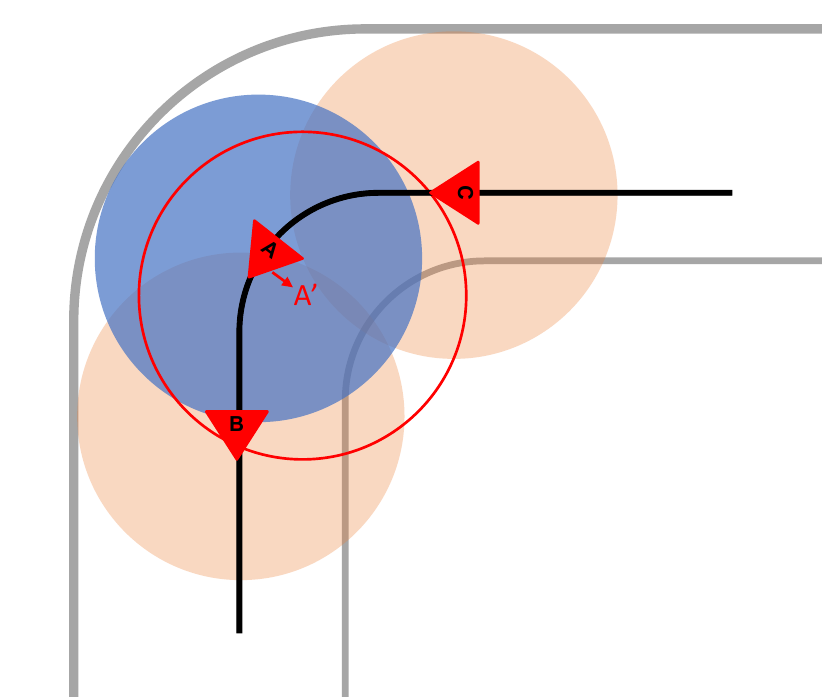}
            \caption{} %
            \label{fig:novel-viewpoint-gen}
            \end{subfigure}
        \end{tabular}
    &
        \begin{tabular}{c}%
        \smallskip
            \begin{subfigure}[t]{0.2\textwidth}
                \centering
                \includegraphics[width=0.9\textwidth]{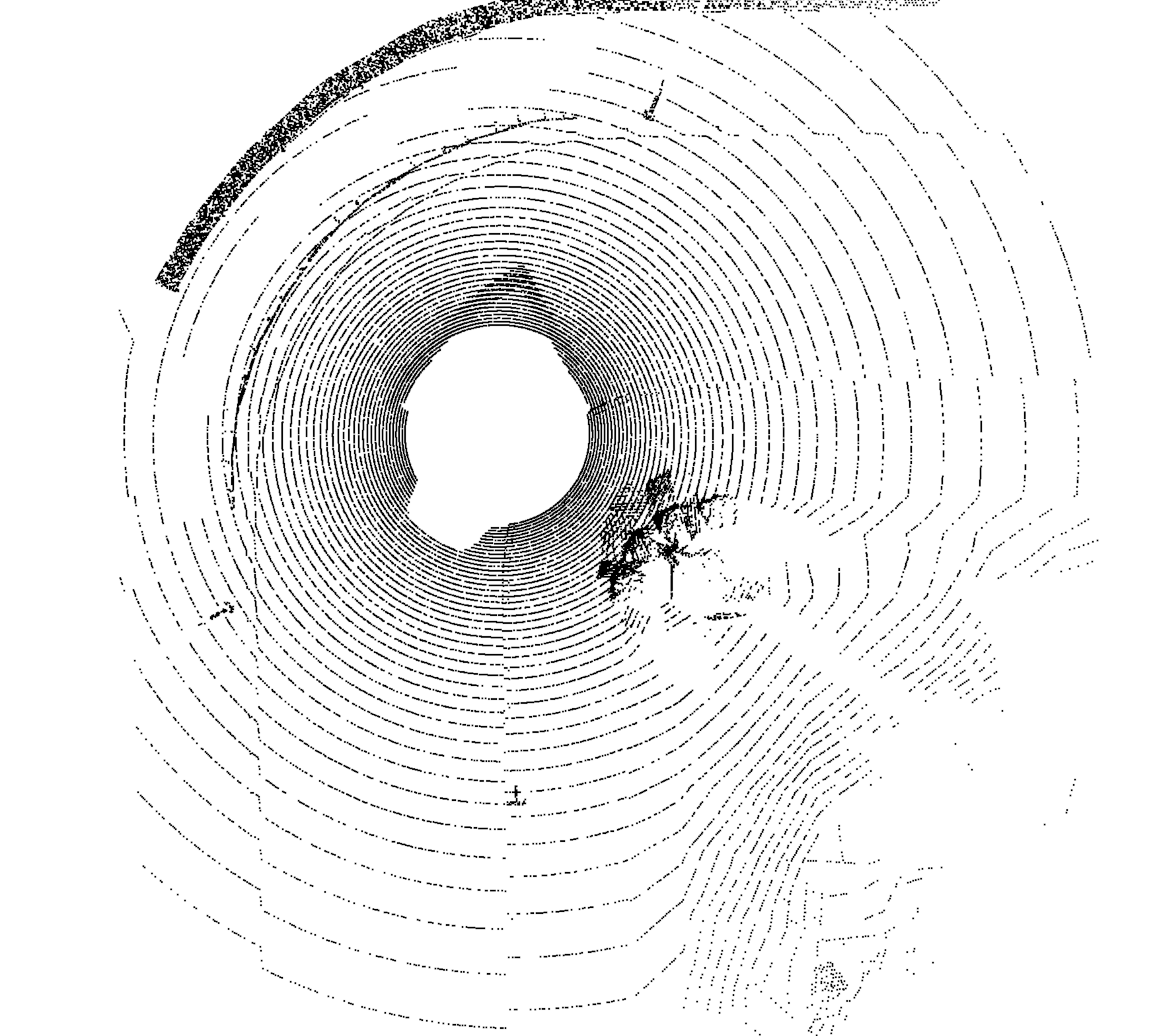}
                \caption{}
                \label{fig:LiDAR-bev}
            \end{subfigure}\\
            \begin{subfigure}[t]{0.2\textwidth}
                \centering
                \includegraphics[width=0.9\textwidth]{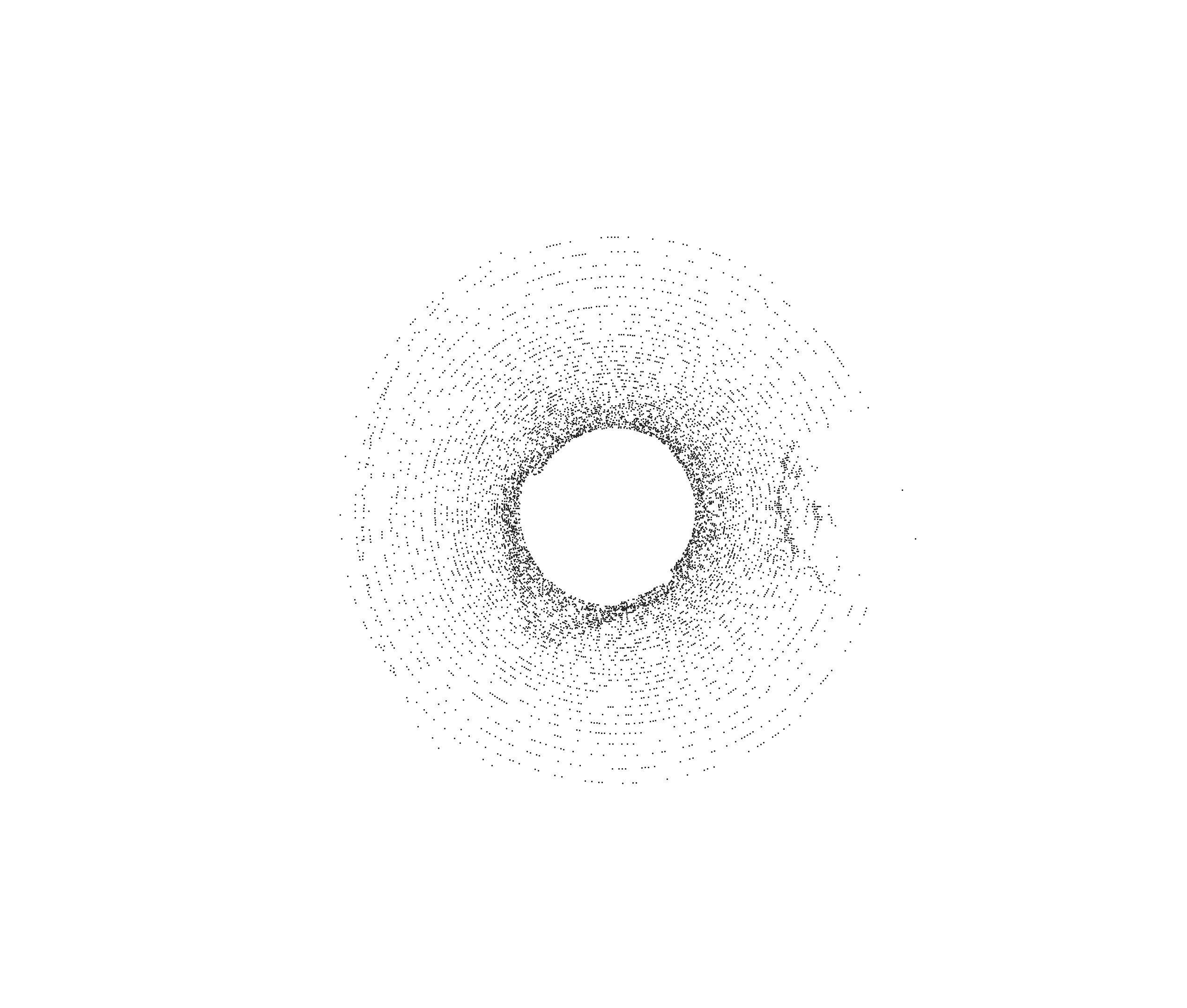}
                \caption{}
                \label{fig:LiDAR-raycast-fail}
            \end{subfigure}
        \end{tabular}
    & 
        \begin{tabular}{c}%
        \smallskip
            \begin{subfigure}[t]{0.2\textwidth}
                \centering
                \includegraphics[width=0.9\textwidth]{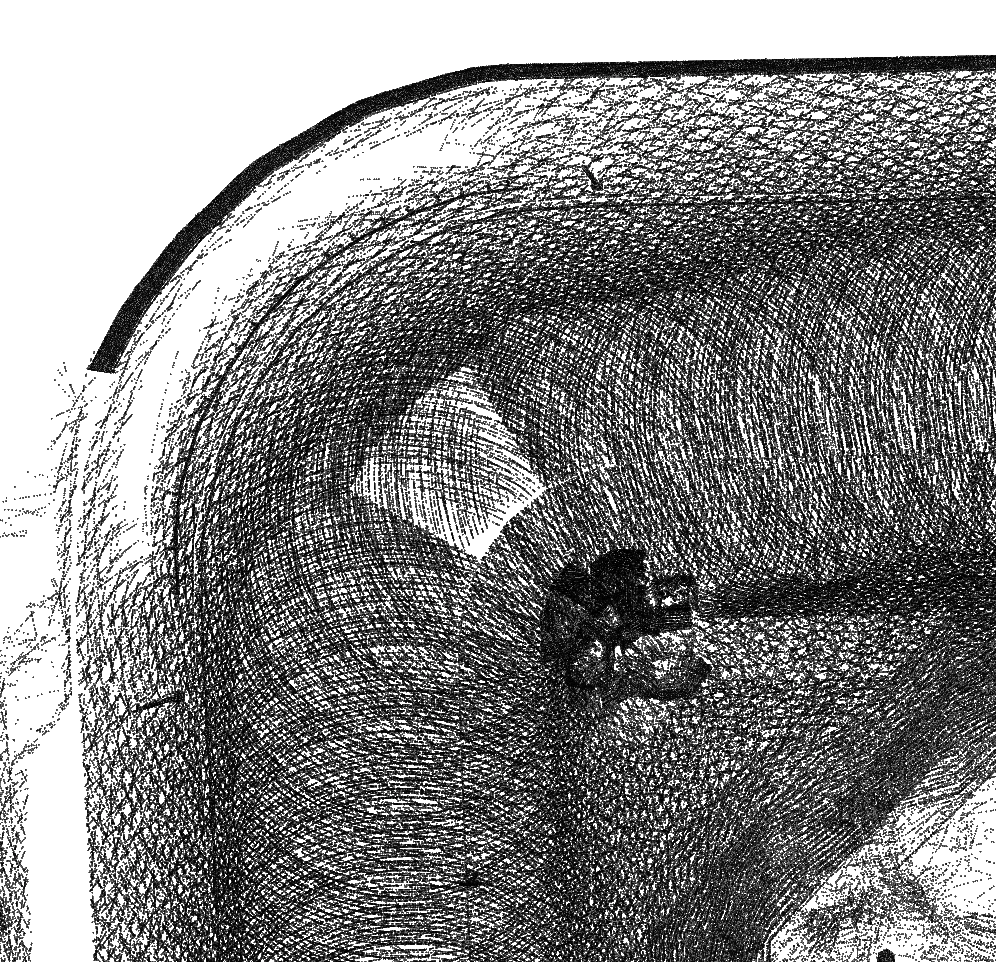}
                \caption{}
                \label{fig:LiDAR-accum}
            \end{subfigure}\\
            \begin{subfigure}[t]{0.2\textwidth}
                \centering
                \includegraphics[width=0.9\textwidth]{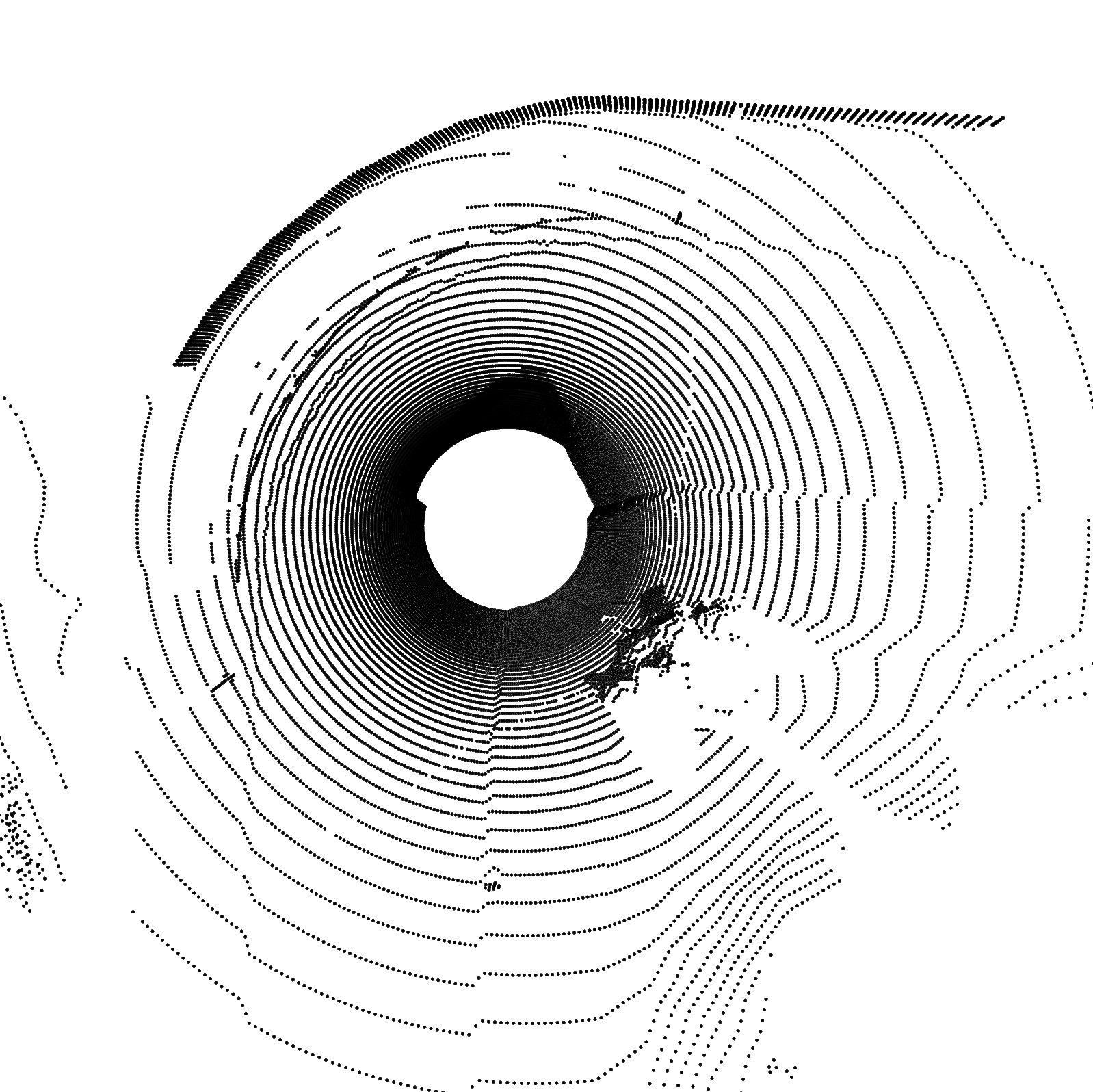}
                \caption{}
                \label{fig:LiDAR-bev-success}
            \end{subfigure}
        \end{tabular}\\
    \end{tabular}
    \caption{The importance of point cloud alignment when synthesizing LiDAR point clouds from novel views: (a) The three circles represent the field of view (FOV) of the LiDAR scans at three positions A, B, and C. We want to synthesize a new point cloud at position A'. A' is lateral to the motion of the vehicle at position A. As can be seen, LiDAR scan A (blue circle) does not entirely cover the FOV of the LiDAR scan at desired position A'(red circle). However, merging point cloud A with point clouds B and C widens the covered area and allows a more comprehensive synthesis of point cloud A'. (b) shows the actual LiDAR at position A. (c) depicts the synthesized point cloud at position A' when using only the LiDAR scan visualized in (b). As can be seen, the synthesized point cloud is relatively sparser and incomplete. (d) shows the result after merging point cloud A with past and future point clouds. (e) shows the synthesized point cloud generated using the entire sequence depicted in (d). This point cloud (e) better represents the ego-vehicle environment. A video providing a supplementary illustration of this can be found \href{https://jonathsch.github.io/lidar-synthesis/\#synthesis}{here}.}
    \label{fig:pc-alignment}
\end{figure*}
   
The next step is accumulating all point clouds from a sequence in the reference coordinate system. Therefore, we use the relative transformations $\mathbf{T}_i$ to project all point clouds $\mathbf{P}_i$ into the coordinate system of $\mathbf{P}_0$:

\begin{equation}
    \mathbf{P'}_i = \mathbf{T}_{i}^{-1} \mathbf{P}_i
\end{equation}

Finally, we simply accumulate the points from all point clouds:

\begin{equation}
    \mathbf{P}^* = \bigcup_{i \in \{0,...,N-1\}}  \{ \mathbf{P'}_i \}
\end{equation}

Figure \ref{fig:pc-alignment} gives a visual description of the importance of this point cloud alignment.

\subsection{Meshing}
\label{sec:meshing}

A LiDAR sensor can be simulated by casting rays into the 3D world and computing the location where they intersect the surface, as done by CARLA~\cite{carla-docs}. However, our scene representation is a point cloud. Its points are infinitesimally small, and the rays are infinitesimally thin. Casting rays into such a point cloud representation would consequently render an empty scene.
Thus, we first reconstruct a triangle mesh from the point cloud. For this, we leverage the Ball Pivoting surface reconstruction algorithm~\cite{bernardini1999ball}, as it preserves the original vertex locations. 

\subsection{Novel-Viewpoint Position Selection}
\label{sec:viewpoint-gen}

Our goal is to train a model for vehicle control, which can recover from errors it made in the past and which caused the vehicle to be located at a lateral offset from the \textit{reference} trajectory. To prepare the model to counteract such scenarios at inference time, we select positions lateral to the reference trajectory where we desire to synthesize novel LiDAR scans. Here we assume that the road topology is locally planar. In Figure \ref{fig:label-generation}a, the red triangles represent the LiDAR sensor poses of the reference trajectory. Meanwhile, the blue triangles are the positions at which the LiDAR scans are desired to be generated.

\subsection{Label Generation}
\label{sec:label-generation}

The next step is to generate future trajectory/waypoint labels for the LiDAR scans in the reference trajectory and at the novel viewpoints. Once this trajectory is known, a waypoint-based controller can be applied to determine the steering and throttle commands.

\begin{figure}
    \centering
    \begin{subfigure}[b]{0.3\linewidth}
    \centering
    \includegraphics[width=\textwidth]{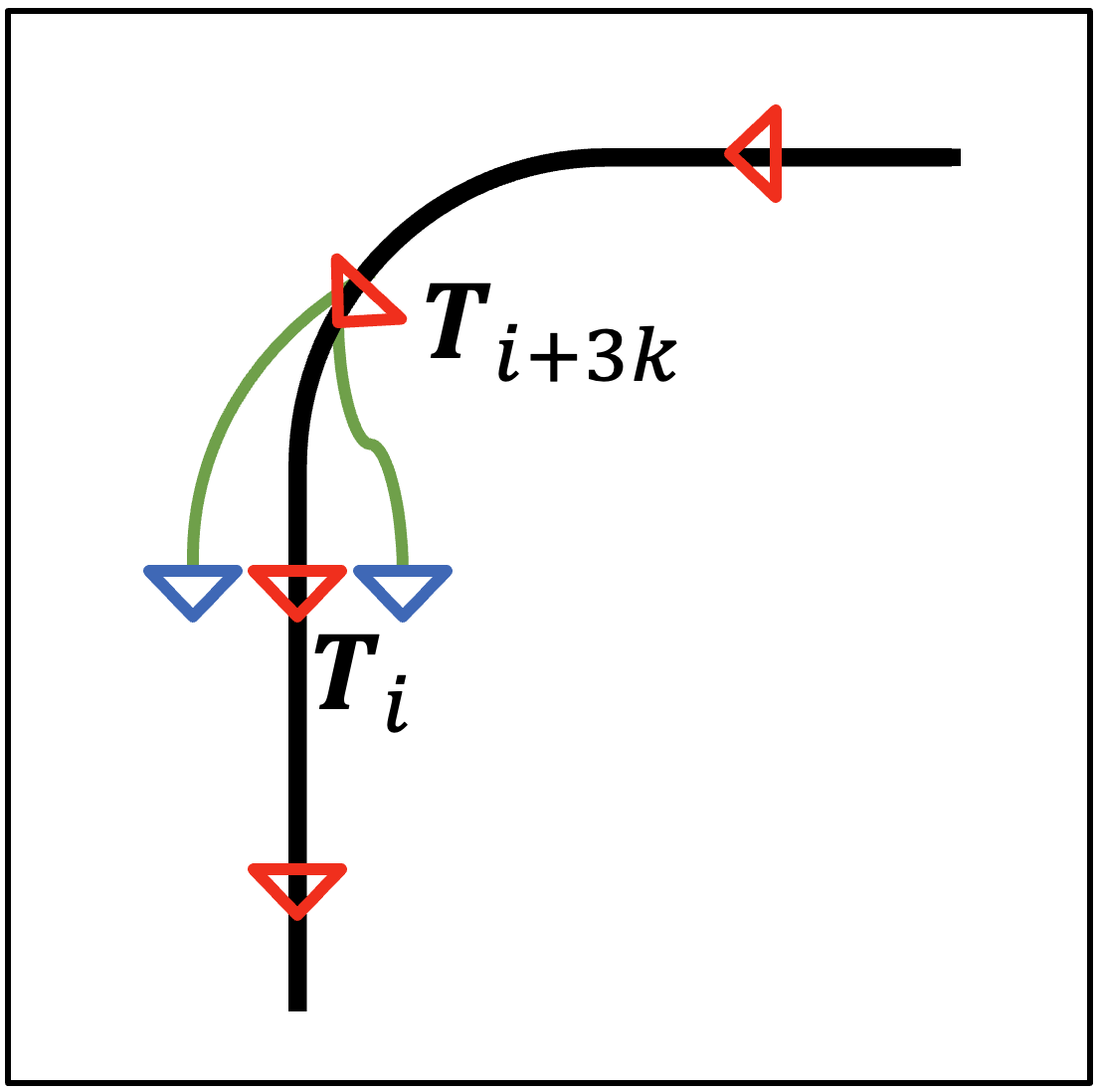}
    \caption{}
    \end{subfigure}
    \begin{subfigure}[b]{0.3\linewidth}
    \centering
    \includegraphics[width=\textwidth]{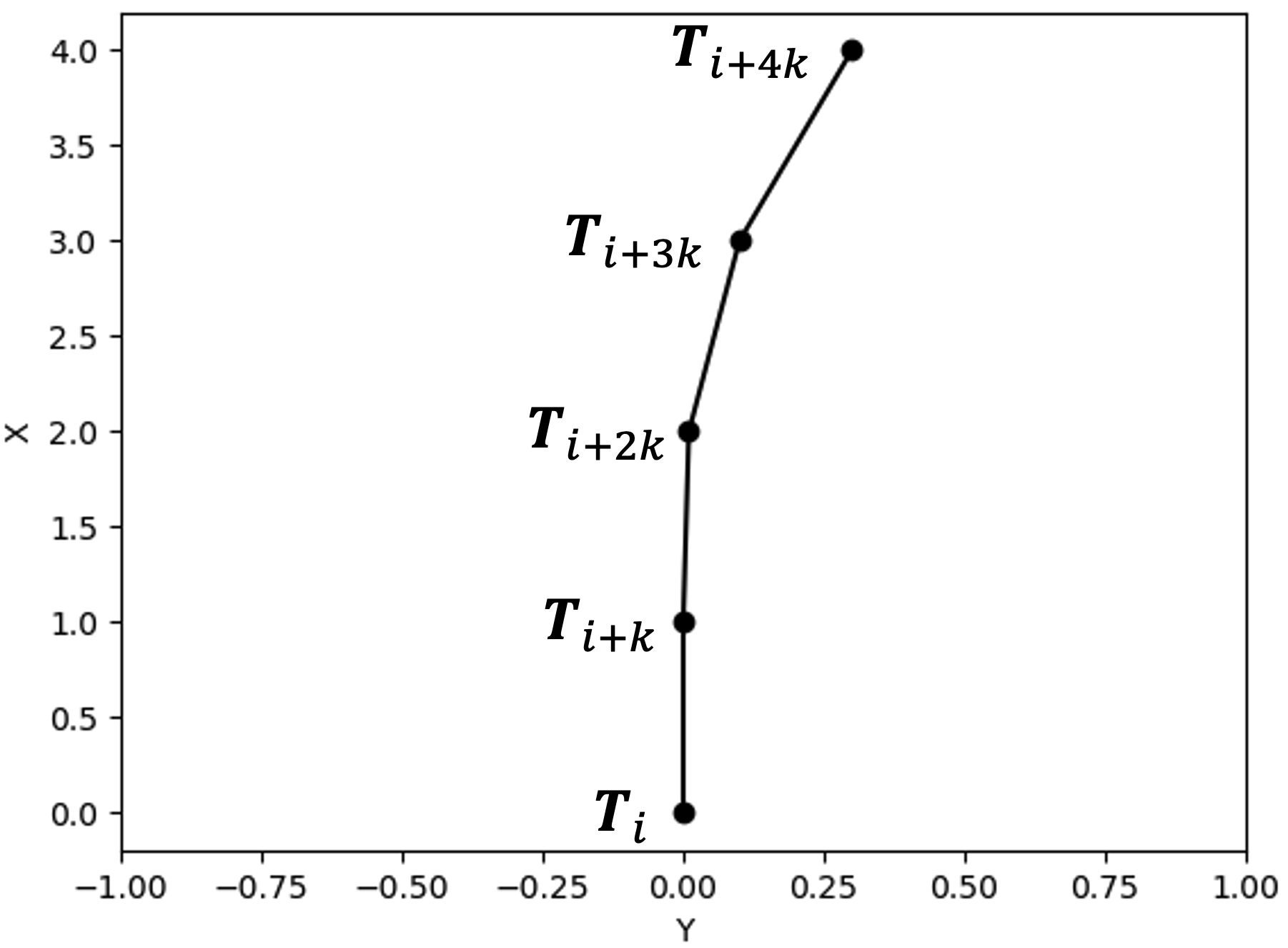}
    \caption{}
    \end{subfigure}
    \begin{subfigure}[b]{0.3\linewidth}
    \centering
    \includegraphics[width=\textwidth]{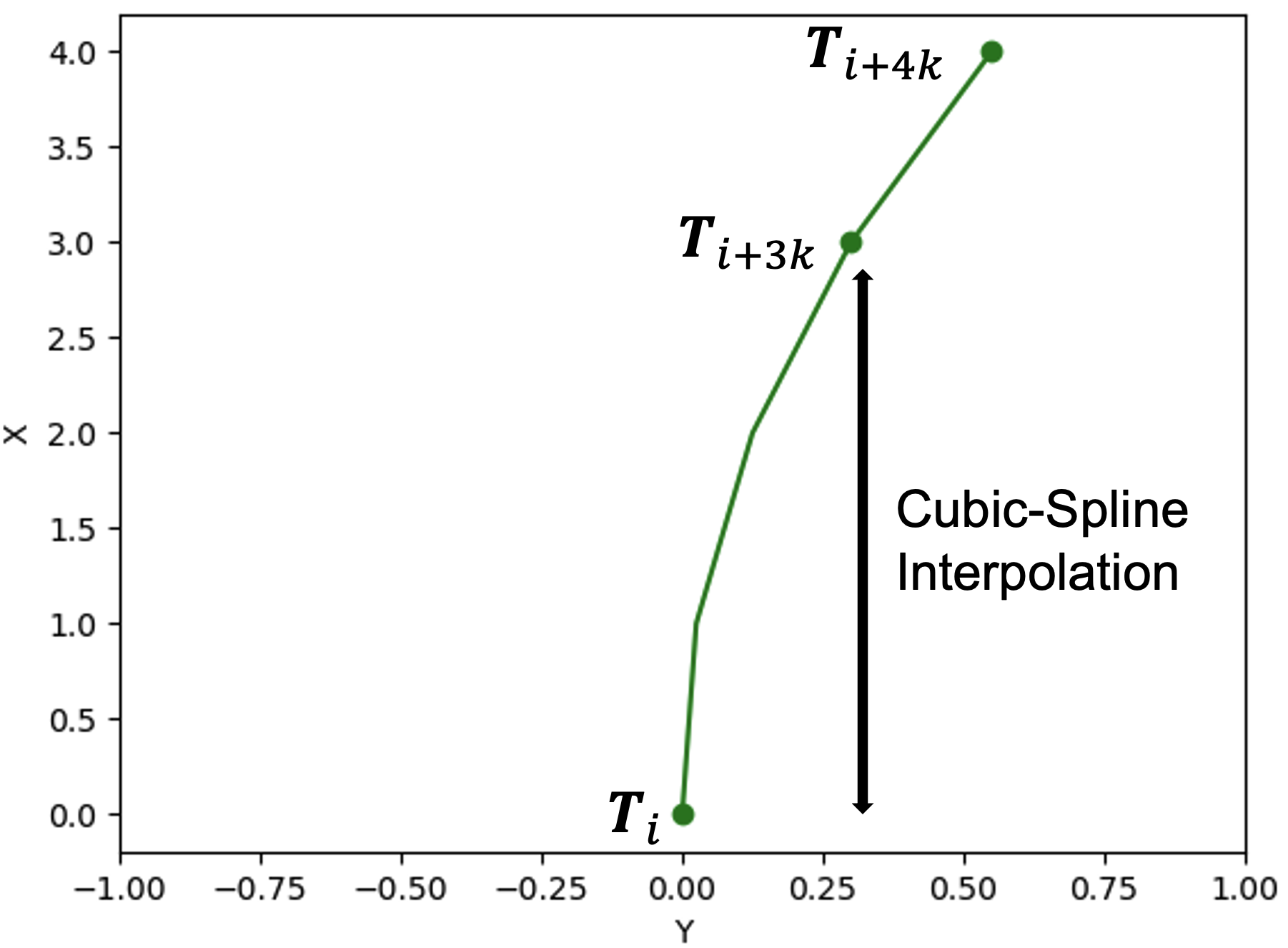}
    \caption{}
    \end{subfigure}
    \caption{(a) depicts the generation of the target trajectory from the current to a future sensor pose. Red poses denote the original sensor locations, blue poses represent the generated, novel positions. The green paths symbolize the computed target trajectory for these novel positions. The target trajectory commences at the pose $T_{i}$ for which the waypoint labels are to be determined. It ends at pose $T_{i+4k}$. $T_{i}$ can either be a position on the reference (b) or a lateral offset (c). Note that $T_{i+3k}$ and  $T_{i+4k}$ will always be the poses on the reference trajectory.}
    \label{fig:label-generation}
\end{figure}

To generate the future waypoints for the reference trajectory point clouds $\mathbf{P}_i$, we transform the poses $\mathbf{T}_{i+k}$, $\mathbf{T}_{i+2k}$, $\mathbf{T}_{i+3k}$, and $\mathbf{T}_{i+4k}$ into the frame of reference of the current frame $\mathbf{T}_{i}$ and consider the x and y component of their translation vector, which represent the 2D waypoints in the coordinate system of $\mathbf{T}_i$. $k$ determines the number of skipped frames when selecting future poses and is found empirically. 
For novel, off-trajectory points, we compute the 2D waypoints of the poses $\mathbf{T}_{i+3k}$ and $\mathbf{T}_{i+4k}$ as described above and compute the remaining waypoints using cubic-spline interpolation, where we set the zeroth waypoint to $(0, 0)$. Figure \ref{fig:label-generation} depicts the schematics of this label-generation process.

 Our deep learning model (\cref{sec:driving-model}) is trained to predict this set of waypoints in the vehicle's coordinate system. It represents the trajectory the car should follow. These waypoints are then fed into a waypoint-based controller, which converts the waypoints into steering, throttle, and brake commands. For this, we adopt the PID controller from \cite{chitta2022transfuser}.

\subsection{Raycasting}
\label{sec:raycasting}

\begin{figure*}
    \centering
    \includegraphics[width=0.8\textwidth]{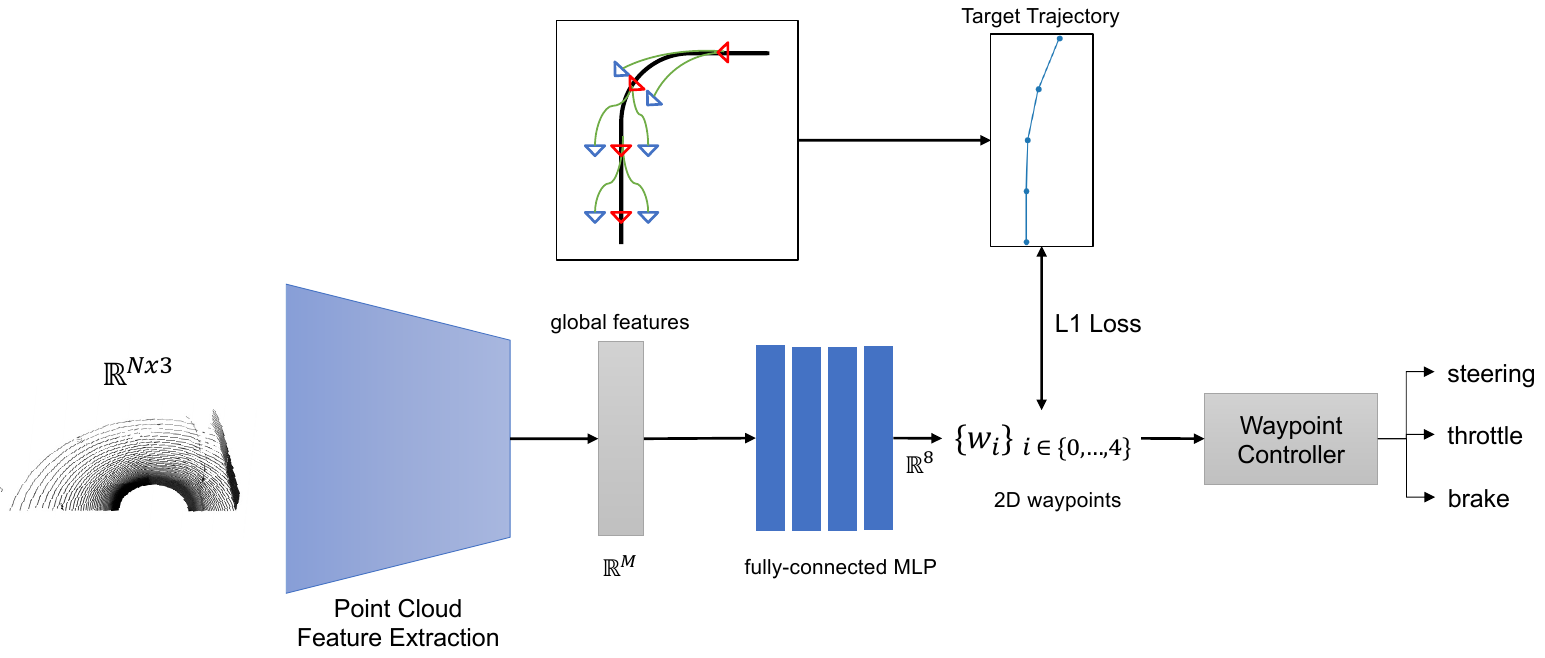}
    \caption{Training and inference of our driving model. The model is trained by minimizing the L1 distance between the predicted and target trajectory represented by a set of 2D waypoints in the vehicle's coordinate frame. These predicted waypoints are mapped to control parameters at inference time using the waypoint-based controller approach from \cite{chitta2022transfuser}.}
    \label{fig:driving-model}
\end{figure*}

For the raycasting, we parameterize rays as a function of scalar $t \in \mathbb{R}$:

\begin{equation}
    \textbf{r}(t) = \textbf{o} + t\textbf{d} \quad 0 \leq t < \infty
\end{equation}

where $\textbf{o} \in \mathbb{R}^3$ is the ray origin, which is solely determined by the desired location of the virtual LiDAR sensor, and $\textbf{d} \in \mathbb{R}^3$ is the normalized ray direction~\cite{pharr2023physically}.

We select the ray directions $\textbf{D} = \{\mathbf{d}_i | i \in 0,...,R-1\}$ to get the desired circular point pattern characteristic for LiDAR scans. To simulate the horizontal rotation for a complete  360\degree~view, angles ($\Phi$) are uniformly sampled between $-\pi$ and $\pi$. The set of rotation matrices $\textbf{H}$, which represents the rotation around the z-axis, is then defined as:

\begin{equation}
    \textbf{H} = \left\{ \begin{pmatrix}cos(\phi) & -sin(\phi) & 0 \\ sin(\phi) & cos(\phi) & 0 \\ 0 & 0 & 1 \end{pmatrix} \mid \phi \in \Phi\ \right\}
\end{equation}

Since a LiDAR sensor has multiple channels (beams), we cast multiple rays per horizontal direction $\textbf{H}$. Hence, we define a fixed set of vertical angles $\Theta$ representing these channels. $\Theta$ heavily depends on the sensor we want to simulate. We evenly sample angles between $\pi/64$ and $\pi/3$ in a logarithmic scale for our experiments. Thereby, we create the set of vectors $\textbf{V}$, representing the vertical ray directions:

\begin{equation}
    \textbf{V} = \left\{ \begin{pmatrix}cos(\theta) \\ 0 \\ sin(\theta) \end{pmatrix} \mid \theta \in \Theta \right\}
\end{equation}

Finally, we can derive the set of directions $\textbf{D}$ as the element-wise rotation of the vertical directions $\vec{v}$ by the horizontal rotations $R$ given by the Cartesian product of \textbf{V} and \textbf{H} as:

\begin{equation}
    \textbf{D} = \left\{ R \vec{v} \mid (R,~\vec{v}) \in \textbf{H} \times \textbf{V} \right\}
\end{equation}

To account for the orientation of the synthesized sensor, we need to multiply all vectors of $\textbf{D}$ with the rotation part of the sensor's transformation matrix.

This framework allows us to simulate different LiDAR hardware by choosing the number and pattern of horizontal and vertical angles $\Phi$ and $\Theta$, respectively. An example of the result is depicted in the bottom-right corner of Figure \ref{fig:method-overview}, where we depict synthesized point clouds at the left-most and right-most lateral offset and on the reference trajectory (at the center).

LiDAR is subject to measurement deviations. We address this by adding Gaussian noise $\mathcal{N}(0, \sigma)$ to the point coordinates and by randomly dropping around 20\% of the points, following the strategy of the simulated LiDAR sensor in CARLA~\cite{Dosovitskiy17carla}.

\subsection{Driving Model}
\label{sec:driving-model}

Our driving model takes a LiDAR point cloud of dimension $N\times3$ as input and predicts a set of 2D waypoints $\{\mathbf{w}_i | \mathbf{w}_i \in \mathbb{R}^2, i \in \{1,...,4\}$ in the $x-y$ plane. Figure \ref{fig:driving-model} depicts the architecture of the model. A 180\degree \space frontal view of the LiDAR point cloud is fed as input, which is most critical when driving forward. It also saves on additional computation compared to the full 360\degree \space view. We use a point cloud feature extraction backbone~\cite{qi2017pointnet, qi2017pointnet++} to output a global feature vector $\in \mathbb{R}^M$. A fully-connected Multi-Layer Perceptron (MLP) then maps from the feature vector to an 8-dimensional output vector, which we reshape into a $2 \times 4$ matrix to observe our four predicted waypoints. Finally, the waypoints are fed into a non-learnable controller, which we adopt from the recent work of~\cite{chitta2022transfuser}. This controller converts the predicted waypoints into steering, throttle, and brake commands. We train the model by minimizing the L1 distance between the predicted trajectory and the target trajectory labels determined from the procedure described in Section \ref{sec:label-generation}.

\section{EXPERIMENTS}\label{sec:experiments}

\subsection{Evaluation Framework}
\label{sec:eval-framework}

The task of vehicle control evaluation falls under the purview of embodied agents, which involve the interaction of an agent with the environment. It has recently emerged as a topic of particular interest to the robotics and autonomous driving community~\cite{anderson2018evaluation}. Numerous datasets exist for testing autonomous driving-related systems \cite{geiger2013kitti, cordts2016cityscapes, caesar2020nuscenes}. However, these datasets only evaluate models in an offline setting, which is inappropriate for assessing the performance of embodied autonomous agents. This is why we evaluate our driving model in an online setup using the open-source driving simulator CARLA~\cite{Dosovitskiy17carla}. It has become a popular platform for benchmarking contemporary autonomous driving algorithms. In fact, we also make comparisons of our model with recent approaches that also conduct their evaluation on CARLA \cite{pccontrol_2022,chitta2022transfuser}. 

The evaluation consists of four routes in the CARLA simulator which we adopt from~\cite{pccontrol_2022} that the agent has to complete without violating any traffic rules. The metrics we use for comparison are 1) The average route completion, 2) The ratio of time which the vehicle remained within its driving lane (\textit{ratio-on-lane})~\cite{control-across-weathers-19}, and 3) The driving score, which is a metric provided by CARLA measuring route completion and traffic rule violations of any kind. Higher values for each of these three metrics are better.

\subsection{Data Collection}
\label{sec:data-collection}

Our proposed framework requires sequences of LiDAR point clouds for training. We collect these point clouds also using CARLA by traversing the ego-vehicle in auto-pilot mode. This is referred to as the \textit{reference trajectory}. It is important to note that no data from the validation routes is collected. We record eight sequences, each consisting of around 300 point clouds.

We then apply our augmentation and label generation pipeline. We use the point-to-point ICP framework KISS-ICP~\cite{vizzo2023kissicp} for pose estimation. For each point cloud, we generate ten additional point clouds at lateral offsets ranging linearly from -2m to +2m, plus target trajectories for every point cloud. 
The vehicle in CARLA can be maneuvered by changing the throttle and steering values. The throttle range is [0,1] while the steering range is [-1, 1], where the maximum value corresponds to 70\degree~for the default vehicle.

\subsection{Models}
\label{sec:exeriments-models}

For comparison, we additionally evaluate the following models:\newline
\noindent\textbf{Baseline (Transfuser):} We use the pre-trained model from ~\cite{chitta2022transfuser} as the baseline. It is a multi-modal (RGB camera + LiDAR) autonomous driving model trained on an order of magnitude larger dataset using ground-truth target trajectories from the CARLA simulator. Thus, Transfuser's results on our evaluation represent the upper bound, and we identify the best model as the one attaining the closest scores to it.

\noindent\textbf{Supervised LiDAR}
This model is trained on the LiDAR point clouds from the collected dataset (see \cref{sec:data-collection}), while the labels are the ground-truth steering commands obtained from the CARLA auto-pilot. The architecture is slightly modified from ours. Instead of predicting the waypoints, the last layer of the MLP is changed to output a scalar steering command instead. This model only predicts the steering angle, so we fix the throttle to 0.4 during inference.

\noindent\textbf{Multi-Camera} follows the approach of~\cite{bojarski2016end}. It is like the supervised LiDAR model, except that data is collected with three RGB cameras and not a single LiDAR sensor. The ground truth steering label used for training is known for the center camera from the CARLA autopilot. A slight bias of +/- 0.2 is added to the steering command for the left and right cameras. We acquire the images on the same routes as our dataset (\cref{sec:data-collection}). During inference, we again fix the throttle to 0.4.

\noindent\textbf{Raycast LiDAR (Ours):} represents the approach trained using the method elaborated in Section \ref{sec:method}. The architecture of this model is explained in Section \ref{sec:driving-model}. We use PointNet++~\cite{qi2017pointnet++} as the feature extraction backbone and a 4-layer MLP with BatchNorm~\cite{pmlr2015batchnorm} and ReLU activations. Note that, unlike all the other models, our approach has no access to the supervised ground truth labels. Rather the labels are inferred from the process described in Section \ref{sec:label-generation}.

\begin{table}
    \centering
    \begin{tabular}{l|c c c}
    \toprule
    
    Method & DS $\uparrow$ & RL $\uparrow$ & RC $\uparrow$ \\
    \midrule
    
    LiDAR (supervised) & 0.39 & 0.67 & 0.59 \\
    
    Multi-Camera~\cite{bojarski2016end} & 0.41 & 0.78 & 0.53 \\
    
    Transfuser~\cite{chitta2022transfuser} & 0.98 & 1.0 & 0.98 \\
    
    Raycast LiDAR (Ours) & 0.90 & 0.93 & 0.97 \\
    
    \midrule
    
    Ablation Study & & & \\
    
    \midrule
    
    Ours (without augmentation) & 0.39 & 0.71 & 0.55 \\
    
    Ours (without alignment) & 0.25 & 0.49 & 0.51 \\ 
    
    \bottomrule
    \end{tabular}
    \caption{Top: Quantitative Comparison of our driving models on the validation routes of our dataset. Bottom: Ablation study showing the impact of skipping point cloud augmentation and point cloud alignment, respectively. DS: Driving Score, RL: Avg. Ratio on Lane, RC: Avg. Route Completion. Higher values for each metric are better.}
    \label{tab:validation-scores}
\end{table}

\begin{table}
    \centering
    \begin{tabular}{l|c c c}
    \toprule
    
    Method & DS $\uparrow$ & RL $\uparrow$ & RC $\uparrow$ \\
    \midrule
    
    Raycast LiDAR (without augmentation) & 0.39 & 0.71 & 0.55 \\
    
    Raycast LiDAR (+- 0.4m) & 0.48 & 0.75 & 0.64 \\
    
    Raycast LiDAR (+- 0.8m) & 0.55 & 0.77 & 0.72 \\
    
    Raycast LiDAR (+- 1.2m) & 0.72 & 0.82 & 0.89 \\
    
    Raycast LiDAR (+- 1.6m) & \textbf{0.90} & \textbf{0.93} & \textbf{0.97} \\
    
    Raycast LiDAR (+- 2.0m) & 0.79 & 0.84 & 0.94 \\
    
    \bottomrule
    \end{tabular}
    \caption{Quantitative Online Evaluation of our driving model trained with increasing lateral offset ranges. DS: Driving Score, RL: Avg. Ratio on Lane, RC: Avg. Route Completion. Higher values for each metric are better.}
    \label{tab:raycast-offset-scores}
\end{table}

\begin{table*}
    \centering
    \begin{tabular}{l|c c c c c c c c c c c c c c | c}
    
    Method & \rotatebox{90}{Clear Noon} & \rotatebox{90}{Cloudy Noon} & \rotatebox{90}{Wet Noon} & \rotatebox{90}{Wet Cloudy Noon} & \rotatebox{90}{Mid Rain Noon} & \rotatebox{90}{Hard Rain Noon} & \rotatebox{90}{Soft Rain Noon} & \rotatebox{90}{Clear Sunset} & \rotatebox{90}{Cloudy Sunset} & \rotatebox{90}{Wet Sunset} & \rotatebox{90}{Wet Cloudy Sunset} & \rotatebox{90}{Mid Rain Sunset} & \rotatebox{90}{Hard Rain Sunset} & \rotatebox{90}{Soft Rain Sunset} & \rotatebox{90}{Average} \\
    \midrule
    Transfuser~\cite{chitta2022transfuser} & 1.0 & 1.0 & 1.0 & 1.0 & 1.0 & 1.0 & 1.0 & 1.0 & 1.0 & 1.0 & 1.0 & 1.0 & 1.0 & 1.0 & 1.0 \\
    Multi-Camera~\cite{bojarski2016end} & 0.78 & 0.77 & 0.74 & 0.73 & 0.65 & 0.61 & 0.66 & 0.69 & 0.72 & 0.59 & 0.62 & 0.58 & 0.57 & 0.60 & 0.66 \\
    RGB-based point cloud\cite{pccontrol_2022} & 0.79 & 0.96 & 0.48 & 0.94 & 0.77 & 0.25 & 0.21 & 0.41 & 0.90 & 0.66 & 0.46 & 0.46 & 0.35 & 0.91 & 0.63 \\
    
    Ours & 0.93 & 0.93 & 0.92 & 0.92 &0.93 & 0.94 & 0.93 & 0.93 & 0.93 & 0.93 & 0.93 & 0.93 & 0.93 & 0.93 & 0.93 \\

    \bottomrule
    \end{tabular}
    \caption{Quantitative comparison of our driving model to concurrent work under different weather conditions. Numbers measure the ratio of the route in which the vehicle remained within its driving lane. A higher value is better.}
    \label{tab:weather-comparision}
\end{table*}

\subsection{Quantitative Results}
\label{sec:quantitative-results}

\begin{figure}
    \centering
    \includegraphics[width=0.9\linewidth]{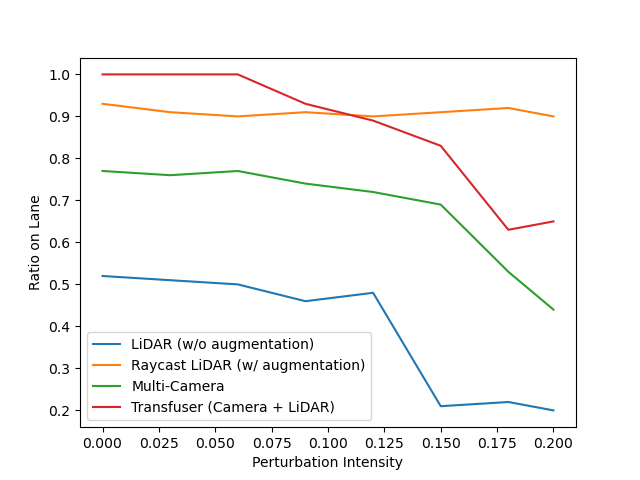}
    \caption{Impact of adding noise to the steering command with increasing intensity, measured by the average ratio the vehicle remains within its driving lane.}
    \label{fig:robustness-perturbation}
\end{figure}

Table, \ref{tab:validation-scores} shows the performance of all models described in the previous Section \ref{sec:exeriments-models}. Note that the performance of our model is closest to the Transfuser baseline model. A qualitative analysis can be found \href{https://jonathsch.github.io/lidar-synthesis/#carla-video}{here}. We also conduct ablation studies to evaluate the influence of synthesizing additional trajectories and aligning the point cloud. It can be noted from the last two rows in Table \ref{tab:validation-scores} that the performance of the model significantly deteriorates when point cloud alignment is not done and when no additional point clouds at off-trajectory positions are synthesized. In Table \ref{tab:raycast-offset-scores}, we further validate the effectiveness of this view synthesis on driving performance. We train multiple models on different subsets of the dataset, gradually increasing the range of generated trajectories from [-0.4m, 0.4m] up to [-2m, 2m].

In Figure \ref{fig:robustness-perturbation}, we test the robustness against perturbations to the steering signal. A zero-mean uniform noise $\mathcal{U}(-\alpha, \alpha)$ of increasing intensity $\alpha$ is sampled. We add the noise to the steering command predicted by the model, with a probability of 0.25. This prevents the noise from canceling out due to its zero mean. This noise causes the vehicle to diverge from the normal trajectory. A model robust to perturbations would be capable of bringing the vehicle back on course. We compare our model against the baseline model without augmentation, the multi-camera model \cite{bojarski2016end}, and the pre-trained Transfuser~\cite{chitta2022transfuser} model. It can be seen that at higher intensity of perturbations, our model is even more stable than the baseline Transfuser model.

In Table \ref{tab:weather-comparision}, we assess the generalization across weather conditions by comparing our model to \cite{bojarski2016end} and \cite{pccontrol_2022}, which use only RGB camera data as input. We also compare with the baseline, i.e., the pre-trained Transfuser~\cite{chitta2022transfuser}, which uses both LiDAR and RGB camera input. As expected, our LiDAR model remains much more robust to weather changes than the RGB models~\cite{bojarski2016end} and \cite{pccontrol_2022}.

\section{DISCUSSION}
\label{sec:discussion}

In this section, we elaborate on the findings from the experimental results and justify why our model performs closest to the baseline Transfuser model.

\noindent{\textbf{No Augmentation:}} The success of our method hinges upon the ability to generate multiple off-trajectory LiDAR scans from a single vehicle traversal. This augmentation diversifies the data seen by the deep learning model during training. Hence, at inference time, if the vehicle diverges from the correct trajectory, the model can bring the vehicle back on track. This is because the model trained with augmented LiDAR data is accustomed to handling such off-trajectory scenarios. In contrast, the supervised LiDAR model trained on the reference trajectory without augmentation shows a significantly worse driving and route completion score, as seen in Table \ref{tab:validation-scores}. This is despite the model having access to the ground truth labels. This is because the supervised model cannot correct the vehicle behavior once the vehicle is off-trajectory.

\noindent{\textbf{No Point Cloud Alignment:}} An essential component of our framework is the alignment of point clouds before proceeding to meshing and ray casting (\cref{sec:alignment}). The bottom row of Table \ref{tab:validation-scores} illustrates what happens if we skip the alignment and reconstruct a triangle mesh from only a single LiDAR point cloud. Due to the sparsity of the point cloud, particularly in areas distant from the sensor, we can only reconstruct a poor-quality mesh from nearby areas, which in turn influences the quality of any synthesized point clouds. We experimentally train another model on point clouds generated without the alignment being applied, the results of which can be found in Table \ref{tab:validation-scores}. We can observe that this model performs much worse compared to when the point clouds were aligned and merged.

\noindent{\textbf{Perturbations:}}
Figure \ref{fig:robustness-perturbation} shows the effect on the model performance when perturbations are added to the vehicle. As can be seen, our model performs consistently across different intensities of perturbations. The driving behavior of all other models sharply deteriorates. Interestingly, even the performance of the baseline Transfuser model drops below our method when the intensity is increased. This is because the Transfuser was not trained on additional off-trajectory data, making it incapable of recovering beyond a specific perturbation limit.
This comparison corroborates how indispensable augmentation is for stable vehicle control. 

\begin{figure}
    \centering
    \includegraphics[width=0.7\linewidth]{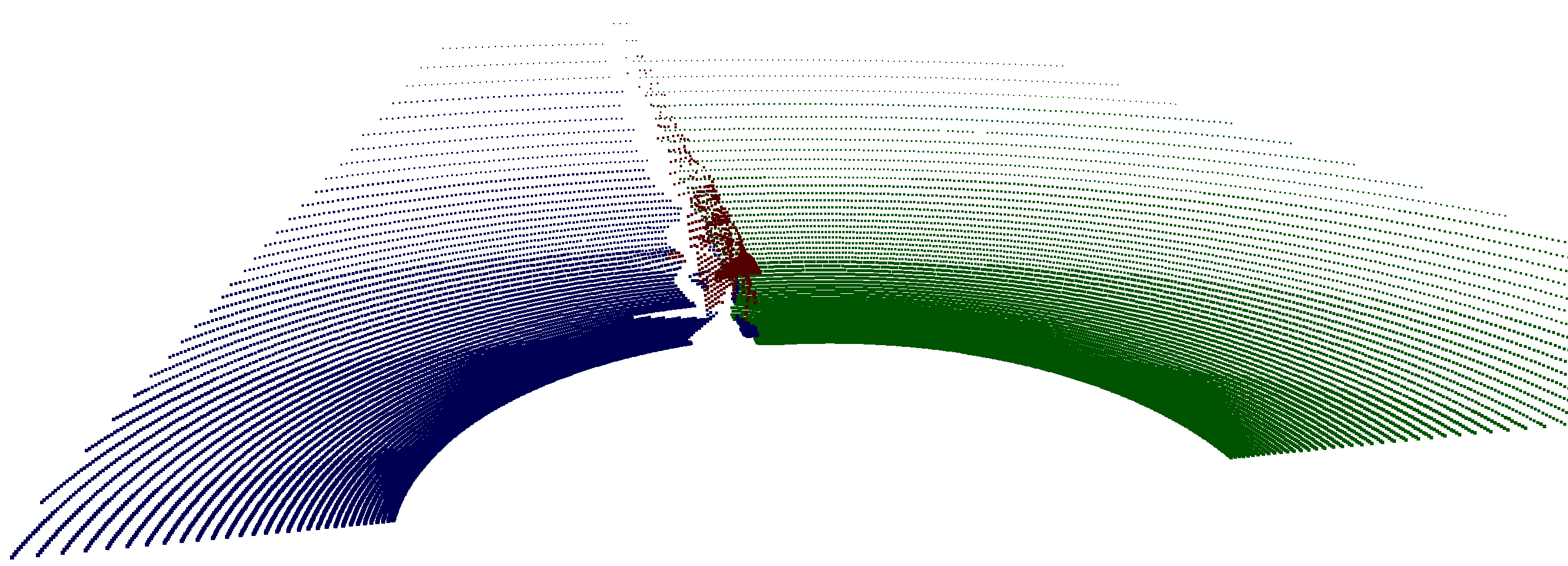}
    \caption{A scenario where the synthesized LiDAR scan lies too far off the reference trajectory. This leads to an unrealistic scenario wherein the car drives straight through the barrier. (The green points denote the road, the blue are the sidewalk, and the red points indicate barriers/fences). An interactive 3D visualization of this can be found \href{https://jonathsch.github.io/lidar-synthesis/\#point-clouds}{here}.}
    \label{fig:raycast-offset-overshoot}
\end{figure}

\noindent{\textbf{Augmentation Range:}}
Table \ref{tab:raycast-offset-scores} shows the performance of six models trained with different ranges of LiDAR augmentation. It can be observed that the performance of a model is enhanced as the range of augmented LiDAR scans is increased up to 1.6 m on either side. This is even beyond the range of the width of the car. The augmentations allow for hallucinating off-trajectory LiDAR traversals at places such as sidewalks and oncoming lanes, which may otherwise be impossible due to safety considerations. Beyond 1.6m, the performance drops slightly. One plausible explanation is that we have barriers and fences at more extreme ranges. This tends to produce unrealistic LiDAR augmentations, such as the car driving midway through the barrier. Figure \ref{fig:raycast-offset-overshoot} depicts one such extreme anomalous scenario.

\noindent{\textbf{Multi-Camera Setup :}}
We have already established the importance of augmenting the dataset with off-trajectory samples. \cite{bojarski2016end} also augment data, but instead of using one, they use three RGB cameras as described in Section \ref{sec:exeriments-models}.
For comparison, we train a model using this three-camera approach. We can see that the performance of such a model is significantly worse than our approach, for which we find two reasons: 1) Our approach can generate an arbitrary number of off-trajectory lateral samples. At the same time, the multi-camera method is limited by the number of cameras placed on the vehicle. 2) The cameras can only be placed within the boundaries of the car's width, which constrains the augmentation range. We have seen in the previous point that the larger the augmentation range, the better the driving performance. Our method is not constrained by the vehicle dimensions and hence, yields better results.

\noindent{\textbf{RGB based Point Cloud Augmentation:}}
To compensate for the limitations of the multi-camera approach, \cite{pccontrol_2022} train a self-supervised depth prediction model from an RGB camera. This predicted depth can then be used to create augmented point clouds. However, the problem with only using RGB images is that they are sensitive to the varying weather/lighting conditions \cite{control-across-weathers-19}. Table \ref{tab:weather-comparision} shows that the performance of the RGB-point cloud model is only good on weather conditions similar to \textit{Cloudy Noon}, on which it was trained. On most other conditions, the performance is unstable and oscillates between high and low driving scores. The same applies to the multi-camera approach since it only uses RGB images. The performance of our method across different weathers is relatively stable across different weather conditions, just like the Transfuser baseline model, which also utilizes LiDAR and was trained on various weather conditions.

\section{CONCLUSION}
\label{sec:conclusion}

We introduced a new method to augment LiDAR point clouds for autonomous driving applications. Using ray-casting, we synthesize LiDAR scans from novel viewpoints and generate corresponding target trajectories without requiring expert driver labels. Our results demonstrate the effectiveness of this approach, particularly in terms of robustness to weather changes and perturbations of the control signal. Although this work focuses on LiDAR-based vehicle control, it can be integrated into a multi-modal driving framework involving additional sensors such as cameras as part of future work.

\bibliographystyle{IEEEtran}
\bibliography{IEEEabrv, egbib}

\end{document}